\documentclass{article}

\usepackage{PRIMEarxiv}

\usepackage[utf8]{inputenc} % allow utf-8 input
\usepackage[T1]{fontenc}    % use 8-bit T1 fonts
\usepackage{hyperref}       % hyperlinks
\usepackage{url}            % simple URL typesetting
\usepackage{booktabs}       % professional-quality tables
\usepackage{amsfonts}       % blackboard math symbols
\usepackage{nicefrac}       % compact symbols for 1/2, etc.
\usepackage{microtype}      % microtypography
\usepackage{lipsum}
\usepackage{fancyhdr}       % header
\usepackage{graphicx}       % graphics
\graphicspath{{media/}}     % organize your images and other figures under media/ folder

\title{Language Diversity: Evaluating Language Usage and AI
Performance on African Languages in Digital Spaces
\thanks{
\textbf{Accepted at Deep Learning Indaba 2026.}} 
}

\author{
  Edward Ajayi, Eudoxie Umwari, Mawuli Deku, Prosper Singadi, Jules Udahemuka \\
  Carnegie Mellon University Africa \\
  Kigali \\
  Rwanda \\
  \texttt{\{eaajayi, eumwari, mdeku, psingadi, judahemuka\}@andrew.cmu.edu} \\
  \And
  Bekalu Tadele \\
  Bahir Dar Institute of Technology \\
  Bahir Dar \\
  Ethiopia \\
  \texttt{btadele@bdu.edu.et} \\
  \And
  Chukuemeka Edeh \\
  Federal University Otuoke \\
  Bayelsa \\
  Nigeria \\
  \texttt{cedeh@fuotuoke.edu.ng} \\
}

\begin{document}
\maketitle

\begin{abstract}
This study examines the digital representation of African languages and the challenges this presents for current language detection tools. We evaluate their performance on Yoruba, Kinyarwanda, and Amharic. While these languages are spoken by millions, their online usage on conversational platforms is often sparse, heavily influenced by English, and not representative of the authentic, monolingual conversations prevalent among native speakers. This lack of readily available authentic data online creates a challenge of scarcity of conversational data for training language models. To investigate this, data was collected from subreddits and
local news sources for each language. The analysis showed a stark contrast between the two sources. Reddit data was minimal and characterized by heavy code-switching. Conversely, local news media offered a robust source of clean, monolingual language data, which also prompted more user engagement in the local language on the news publishers’ social media pages. Language detection models, including a macro-classifier (GlotLID), the specialized AfroLID, and a general-purpose LLM (Llama 3.3 70B), performed with near-perfect accuracy on the clean news data but struggled with the code-switched Reddit posts. The study concludes that professionally curated news content is a more reliable and effective source for training context-rich AI models for African languages than data from conversational
platforms. It also highlights the need for future models that can process clean and code-switched text to improve the detection accuracy for African languages.
\end{abstract}

\section{Introduction}
The African continent is characterized by immense linguistic diversity, with over 2,000 languages spoken by a population of more than 1.5 billion people \cite{sciencedirect_african_languages}. This rich tapestry of languages facilitates cultural exchange and communication across a vast array of communities. However, the representation of African languages in the digital sphere has not kept pace with advancements in technology. While language technologies, particularly Large Language Models (LLMs), have demonstrated remarkable capabilities for widely-spoken global languages, their performance on African languages remains a significant challenge \cite{alhanai2025bridging}.
\noindent
To investigate this issue, we evaluate the digital presence of African languages on conversational platforms and assess the performance of multiple language identification tools (spanning a general macro-classifier, a specialist African LID model, and a general-purpose LLM) in identifying them. Our investigation focuses on three prominent languages:

\begin{itemize}
    \item \textbf{Yoruba:} A highly tonal, analytic language belonging to the Volta-Niger branch of the Niger-Congo language family \cite{ethnologue_yoruba}. With over 50 million speakers primarily in Nigeria, Benin, and Togo \cite{wiki_yoruba}, it utilizes a Latin-based script with extensive diacritics. It represents a high-resource language within the African context, yet exhibits unique complexities for language models due to its strict tonal morphology.
    \item \textbf{Kinyarwanda:} A heavily agglutinative Bantu language belonging to the broader Niger-Congo family \cite{ethnologue_kinyarwanda}. Spoken by approximately 12 million people primarily in Rwanda \cite{wiki_kinyarwanda}, it utilizes a standard Latin script without tonal diacritics in its common digital orthography. It provides a structural contrast to Yoruba, representing the complex noun-class systems typical of the vast Bantu language group.
    \item \textbf{Amharic:} The official working language of Ethiopia's federal government and the second-most spoken Semitic language globally after Arabic \cite{wiki_amharic}. Crucially, it belongs to the Afroasiatic language family \cite{ethnologue_amharic} and is natively written in the Ge'ez (Ethiopic) script. This provides a vital orthographic and phylogenetic contrast to the Latin-script Niger-Congo languages, allowing us to evaluate how script boundaries impact AI detection systems.
\end{itemize}

\section{Problem Statement}

Despite having millions of speakers, the digital footprint of these languages on online platforms is notably limited. Our empirical analysis reveals that the use of Yoruba, Amharic, and Kinyarwanda is often sparse and not representative of natural, real-world conversation. On X (formerly Twitter), we observed infrequent conversational use of these languages. On Reddit, the situation is similar; for instance, the Yoruba subreddit primarily features requests for English-to-Yoruba translations or questions heavily code-switched with English, rather than authentic discourse \cite{reddit_yoruba}. This pattern of English dominance is also prevalent on subreddits for Amharic \cite{reddit_amharic} and Kinyarwanda \cite{reddit_rwanda}.
\noindent
This low and unrepresentative digital presence creates a two-fold problem for both linguistic communities and technology development:

\begin{itemize}
    \item \textbf{Scarcity of Authentic Training Data:} The lack of robust, conversational data in these languages online hinders the development and fine-tuning of effective language technologies. LLMs trained on existing web data are often ill-equipped to understand and generate these languages as they are used colloquially.
    \item \textbf{A Negative Feedback Loop:} The limited and skewed data available online leads to poor performance of AI models in tasks like language identification and translation. This poor performance can, in turn, discourage users from using their native languages on digital platforms, further exacerbating the data scarcity problem.
\end{itemize}

\section{Related Work}

The challenge of representing African languages in the digital age has been approached from several perspectives. One major area of focus has been the creation of datasets to improve NLP model performance. For instance, \cite{alhanai2025bridging} developed a dataset of approximately one million human-translated sentences in eight African languages to fine-tune LLMs, demonstrating significant performance gains. Similarly, the Masakhane initiative organized NLP practitioners to build a named-entity recognition (NER) dataset for ten African languages, highlighting transfer learning potential and open research challenges \cite{adelani_masakhaner_2021}. Recognizing the foundational need for structured data, \cite{magangane_datasets_2024} developed a framework for systematic data collection and documentation for South African languages, which they used to build a language identification model. On the language identification front, \cite{kargaran2023glotlid} introduced GlotLID, a fastText-based macro-classifier covering over 1,600 language varieties, including low-resource African languages, providing a strong general-purpose LID baseline. Complementarily, AfroLID \cite{afrolid_acl} offers a specialized neural LID system covering 517 African languages and varieties. These two tools represent distinct design philosophies (breadth versus depth), making their comparative evaluation on the same social media corpus a meaningful contribution to the field.
\noindent
A parallel body of research examines the role of new media in language preservation and endangerment. Language is a complex support system essential for social function, and modern digital platforms offer new avenues for its documentation and revitalization \cite{nezvitskaya_protecting_2021}. Some scholars are optimistic; \cite{moseley_endangered_2023} suggests that platforms such as YouTube can enhance digital linguistic diversity by providing low-resource languages with broader reach through accessible video and music content. This aligns with the findings that social media platforms like blogs can be ideal spaces for cultural and intercultural learning, even if not specifically focused on African languages \cite{mediause2019}.
\noindent
However, others express caution, arguing that the same forces of globalization and urbanization that threaten languages offline are amplified in the digital realm. \cite{xolmatova_endangered_2025} contends that the widespread use of dominant languages such as English and Mandarin contributes to the decline of indigenous languages, as younger generations often adopt languages that appear to offer greater economic and social opportunities.
\noindent
Building on this existing research, our study makes three concrete contributions: (1) we provide a large-scale, multi-year empirical analysis of authentic African language use on Reddit versus professionally curated news media; (2) we conduct a direct comparative evaluation of three LID systems (GlotLID, AfroLID, and Llama 3.3 70B) on the same corpora; and (3) we quantify code-mixing patterns using a Code-Mixing Index (CMI) to explain the performance gap between clean and conversational data.
\section{Methodology}
\begin{figure}[htbp]
    \centering
    \includegraphics[width=0.9\linewidth]{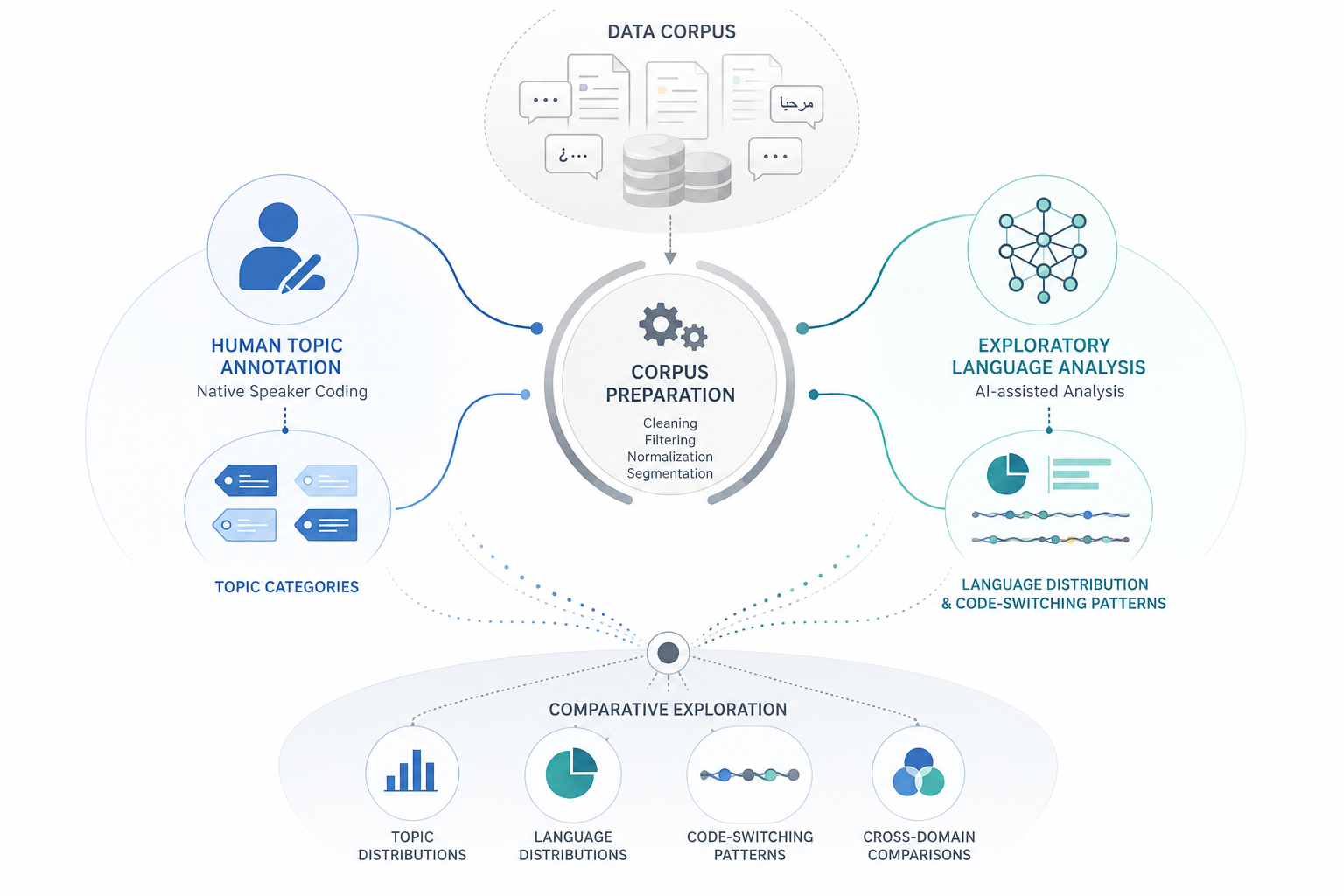}
    \caption{Workflow diagram of the study’s methodology, detailing the progression from data corpus preparation to comparative exploration.}
    \label{fig:methodology}
\end{figure}

The methodology followed in this study is illustrated in Figure~\ref{fig:methodology}. The process begins with the compilation of a data corpus consisting of text collected from social media and professional news platforms. This raw data undergoes rigorous corpus preparation, including cleaning, filtering, and normalization, to ensure high-quality inputs. The prepared corpus is then split into two parallel analytical tracks. The first track, human topic annotation, utilizes native speaker coding to categorize the texts into distinct topic categories. The second track, exploratory language analysis, employs AI-assisted tools to determine language distributions and code-switching patterns. Finally, the outputs from both tracks converge into a comparative exploration phase, allowing us to empirically contrast topic distributions, language usage, and code-switching patterns across different digital domains.

\subsection{Data Sourcing}
For this project, we extracted data from two different sources to understand how people interact and the language concentration on different platforms.
\subsubsection{Reddit}
Due to the infrequent use of social media platforms like Reddit compared to Twitter or Facebook, we scraped three subreddits: r/yoruba\cite{reddit_yoruba}, r/amharic\cite{reddit_amharic}, and r/Rwanda\cite{reddit_rwanda}. These subreddits correspond to the Yoruba, Amharic, and Kinyarwanda languages, respectively. We used PRAW\cite{geeksforgeeks_reddit_scraping} to collect all available posts spanning multiple years from each of these communities, providing a broad temporal coverage of conversational language use.
\subsubsection{News Pages}
Because English was the most dominant language in the scraped Reddit data, we sought news channels that post primarily in local languages. For Yoruba, we scraped BBC Yoruba\cite{bbc_yoruba}. For Kinyarwanda, we used Rwanda Broadcasting Agency\cite{rba_rwanda}, and for Amharic, we used Fana Broadcasting Corporation\cite{fana_media}. This approach provided a robust concentration of local languages. We also examined the social media pages of these news channels, where we found numerous comments in local languages on their Facebook posts. 
\subsection{Data Preprocessing}
To ensure the integrity of the language identification process, a rigorous text cleaning pipeline was applied to the raw scraped data. The preprocessing steps included: (1) removing missing (NaN) or empty entries; (2) using regular expressions to strip all URLs; (3) removing emojis, handles, and irregular special characters while preserving alphanumeric characters and basic punctuation; and (4) collapsing multiple consecutive whitespaces into a single space. Finally, any post containing 10 characters or fewer after this cleaning process was excluded from the analysis to prevent unreliable language detection on fragments too short to provide meaningful linguistic context.

\subsection{Language Detection}
In this research, we employed three complementary language identification tools to handle the linguistic diversity and potential code-switching present in our datasets. First, we used GlotLID\cite{kargaran2023glotlid}, a fastText-based macro-classifier supporting over 1,600 language varieties, as a general-purpose baseline. Second, we utilized AfroLID\cite{afrolid_acl}, a specialized neural language identification framework designed specifically for African languages, covering 517 languages and varieties. Third, because AfroLID does not support English detection and we anticipated significant English code-switching in the social media data, we incorporated Llama 3.3 70B, an open-source multilingual large language model capable of identifying both African languages and English within code-switched contexts. To evaluate the LLM, it was prompted in a zero-shot setting with the instruction: \textit{"You are a language detection expert. Analyze the following text and determine if the primary language is English, Yoruba, Kinyarwanda, or Amharic."} To handle ambiguity, the prompt explicitly instructed the model to output the dominant language in cases of code-switching, and to output 'Unknown' if the language could not be determined. This three-way evaluation design allows us to compare a broad macro-classifier, a specialist African LID model, and a general-purpose LLM on the same corpora.
\subsection{Topic Classification}
To understand the context of what constitutes the most common news topics in these languages, we conducted a topic classification of the news articles. For this process, four native speakers (one for Yoruba, one for Amharic, and two for Kinyarwanda) annotated the titles by topic. The categories used for annotation included Business, Education, Sports, and History, among others. To maximize data coverage, the datasets were partitioned such that each article was labeled by a single native speaker without overlap; consequently, inter-annotator agreement (IAA) metrics are not applicable. This manual annotation process provided a robust understanding of the kinds of conversations and topics that are prominent in the news media for each language.
\section{Results and Discussion}

Our analysis reveals significant patterns in the digital representation of Yoruba, Kinyarwanda, and Amharic languages across different platforms and the varying performance of language detection tools on these datasets. The findings are categorized into three key areas: digital language usage on social media, a comparison of language detection models, and an analysis of prominent topics in news media.
\subsection{Language Usage on Social Media}
Our investigation into the usage of African languages on social media platforms revealed a pervasive trend of English dominance and code-switching. A qualitative analysis of the scraped Reddit data indicates that when these languages are used, they are often not in a conversational context. Instead, their usage is limited, primarily appearing in questions or for translation purposes, with a heavy prevalence of English. This observation aligns with the relatively low total number of posts collected from the subreddits over the expansive five-year longitudinal window, with Yoruba yielding 629 posts, Amharic 465 posts, and Kinyarwanda 994 posts.

\begin{figure}[htbp]
    \centering
    \includegraphics[width=0.48\linewidth]{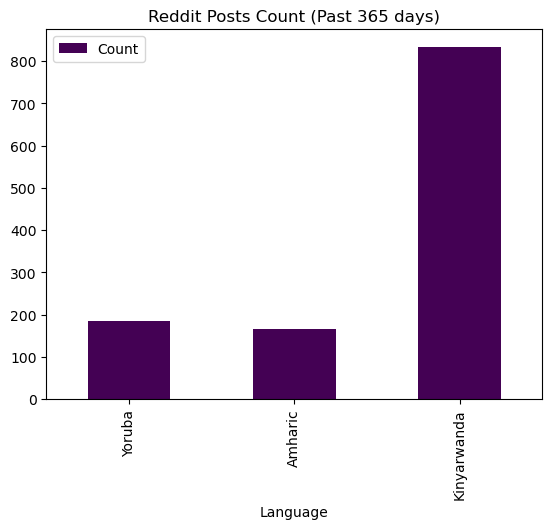}\hfill
    \includegraphics[width=0.48\linewidth]{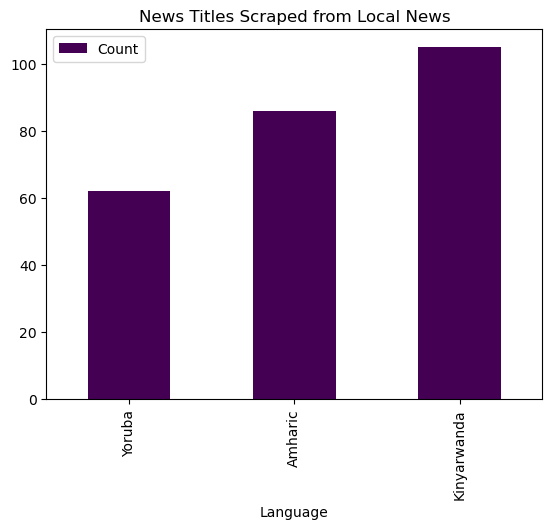}
    \caption{Comparison of Reddit Posts and News Articles Count. Left: Number of Reddit Posts. Right: Number of News Articles.}
    \label{fig:combined_counts}
\end{figure}

\begin{figure}[htbp]
    \centering
    \includegraphics[width=0.48\linewidth]{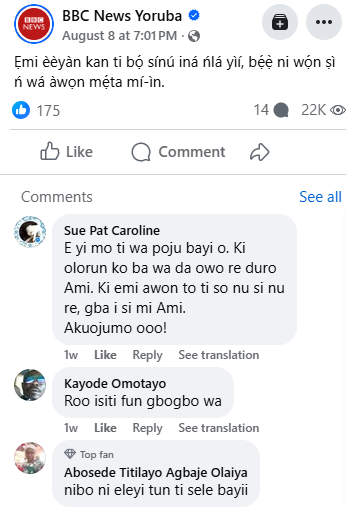}\hfill
    \includegraphics[width=0.48\linewidth]{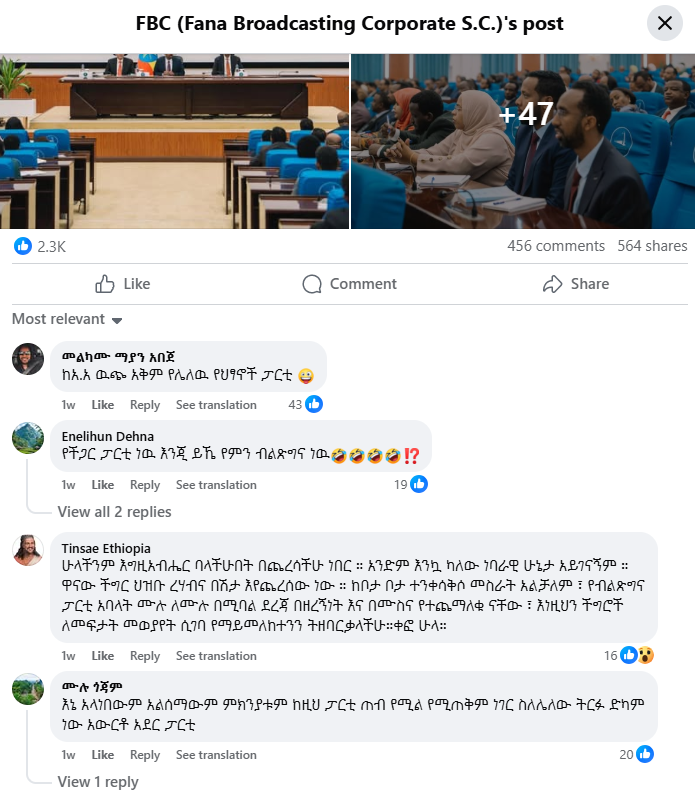}
    \caption{Comparison of BBC Yoruba and FANA Broadcasting Comments. Left: BBC Yoruba Comments. Right: FANA Broadcasting Comments.}
    \label{fig:combined_comments}
\end{figure}

\noindent
Given this sparsity and the code-switched nature of the Reddit data, we sought an alternative, more robust source of language data. We found a different dynamic on the social media pages of local news channels. When a post is initiated in a local African language on platforms like Facebook, it elicits a significant percentage of responses in the same local language as seen in Figure \ref{fig:combined_comments}. This demonstrates that content generated in the local language, without code-switching, can stimulate authentic conversational engagement. Building on this observation, the actual news articles and titles scraped from these professional sources yielded a highly concentrated repository of clean, non-code-switched language data compared to Reddit, providing a robust evaluation dataset of 62 articles in Yoruba, 86 in Amharic, and 105 in Kinyarwanda.

\subsection{Language Detection Performance}
Our language detection analysis provides quantitative support for the qualitative observations we made about language usage. By comparing the results of a fastText-based macro-classifier (GlotLID), a specialized African language framework (AfroLID), and a state-of-the-art multilingual LLM (Llama 3.3 70B), we gain a clearer picture of the data's composition and the models' respective strengths and weaknesses. To account for the variance in dataset sizes, all reported proportions are accompanied by 95\% Wilson Score Intervals\cite{wilson1927probable}, calculated as:
\begin{equation}
    p \approx \frac{\hat{p} + \frac{z^2}{2n}}{1 + \frac{z^2}{n}} \pm \frac{z}{1 + \frac{z^2}{n}} \sqrt{\frac{\hat{p}(1-\hat{p})}{n} + \frac{z^2}{4n^2}}
\end{equation}
where $\hat{p}$ is the observed proportion, $n$ is the total sample size, and $z \approx 1.96$ is the critical value.
\subsubsection{Performance on Social Media (Reddit)}
As shown in Table \ref{tab:reddit_detection}, all three models struggled to identify the target African languages on Reddit, revealing a complex, code-switched linguistic environment. Llama 3.3 70B predominantly classified the posts as English (e.g., 97.6\% for \texttt{r/Rwanda}), reflecting the true dominance of English vocabulary in the dataset. While AfroLID correctly identified Kinyarwanda in 62.7\% of the posts on \texttt{r/Rwanda}, it hallucinated other African languages (such as Swahili and Wolof) for the heavily English-mixed text since it cannot output English. GlotLID, forced to choose from its 1,600 language classes, primarily predicted English but suffered very low native classification rates (e.g., 2.3\% for \texttt{r/Rwanda}). 

\begin{table}[htbp]
\centering
\caption{Language Detection on Conversational Reddit Data (5-Year). The primary percentages indicate the proportion of total posts ($n$) identified as the target native language by each classifier. The values in brackets represent the 95\% Wilson Score Confidence Intervals, indicating the statistical margin of error. Llama 3.3's classification of English is included for context to demonstrate the prevalence of code-switching.}
\begin{tabular}{lccc}
\toprule
\textbf{Metric} & \textbf{r/Yoruba} & \textbf{r/Rwanda} & \textbf{r/Amharic} \\
\midrule
Posts ($n$) & 629 & 994 & 465 \\
GlotLID (Native) & 21.6\% [18.6-25.0] & 2.3\% [1.5-3.4] & 7.5\% [5.5-10.3] \\
AfroLID (Native) & 35.8\% [32.1-39.6] & 62.7\% [59.6-65.6] & 22.4\% [18.8-26.4] \\
Llama 3.3 (Native) & 8.4\% [6.5-10.9] & 2.0\% [1.3-3.1] & 15.7\% [12.7-19.3] \\
Llama 3.3 (English) & 90.5\% & 97.6\% & 84.3\% \\
\bottomrule
\end{tabular}
\label{tab:reddit_detection}
\end{table}

\subsubsection{Performance on News Media}
In stark contrast, when applied to news media, all three systems demonstrated near-perfect accuracy. As detailed in Table \ref{tab:news_detection}, Llama 3.3 70B correctly identified 100\% of the articles across all three languages. AfroLID and GlotLID also achieved near-perfect performance. This highlights the effectiveness of all three tools on clean, non-code-switched text, confirming the value of news media as a robust source for training and evaluation data.

\begin{table}[htbp]
\centering
\caption{Language Detection on Structured News Media. The primary percentages indicate the proportion of total articles ($n$) correctly identified as the target native language. The values in brackets represent the 95\% Wilson Score Confidence Intervals, establishing the statistical reliability of the models' near-perfect performance on these smaller, clean datasets.}
\begin{tabular}{lccc}
\toprule
\textbf{Metric} & \textbf{BBC Yoruba} & \textbf{Kinyarwanda} & \textbf{Fana Amharic} \\
\midrule
Articles ($n$) & 62 & 105 & 86 \\
GlotLID (Native) & 100.0\% [94.2-100] & 91.4\% [84.5-95.4] & 98.8\% [93.7-99.8] \\
AfroLID (Native) & 100.0\% [94.2-100] & 97.1\% [91.9-99.0] & 100.0\% [95.7-100] \\
Llama 3.3 (Native) & 100.0\% [94.2-100] & 100.0\% [96.5-100] & 100.0\% [95.7-100] \\
\bottomrule
\end{tabular}
\label{tab:news_detection}
\end{table}

\subsection{Code-Mixing Index (CMI) Analysis}
To empirically quantify the linguistic composition causing the models to fail on Reddit, we calculated a Code-Mixing Index (CMI) for each post. Inspired by standard code-mixing metrics\cite{gamback2014measuring}, we formulate two dataset-specific metrics. For the Latin-script languages (Yoruba and Kinyarwanda), we define CMI as the fraction of tokens in the post that intersect with a 10,000-word English dictionary $D_{eng}$:
\begin{equation}
    CMI_{Latin} = \frac{| \{ t \in T \mid t \in D_{eng} \} |}{|T|}
\end{equation}
where $T$ represents the set of all tokens in the post. For Amharic, we define CMI computationally as the fraction of Latin characters relative to the total valid character count:
\begin{equation}
    CMI_{Amharic} = \frac{C_{Latin}}{C_{Latin} + C_{Ethiopic}}
\end{equation}
where $C$ represents the respective character counts. Under both formulations, a CMI of 0.0 indicates a pure native text, while 1.0 indicates a text entirely in English. Because Amharic natively utilizes the Ge'ez script, the high concentration of Latin characters could theoretically indicate Romanized Amharic transliteration. However, a qualitative review of the high-CMI Amharic posts confirms they are predominantly English sentences (e.g., users asking for translations or discussing the language in English), thereby validating the Latin character ratio as a reliable proxy for English code-mixing in this dataset.

\begin{figure}[htbp]
    \centering
    \includegraphics[width=0.48\linewidth]{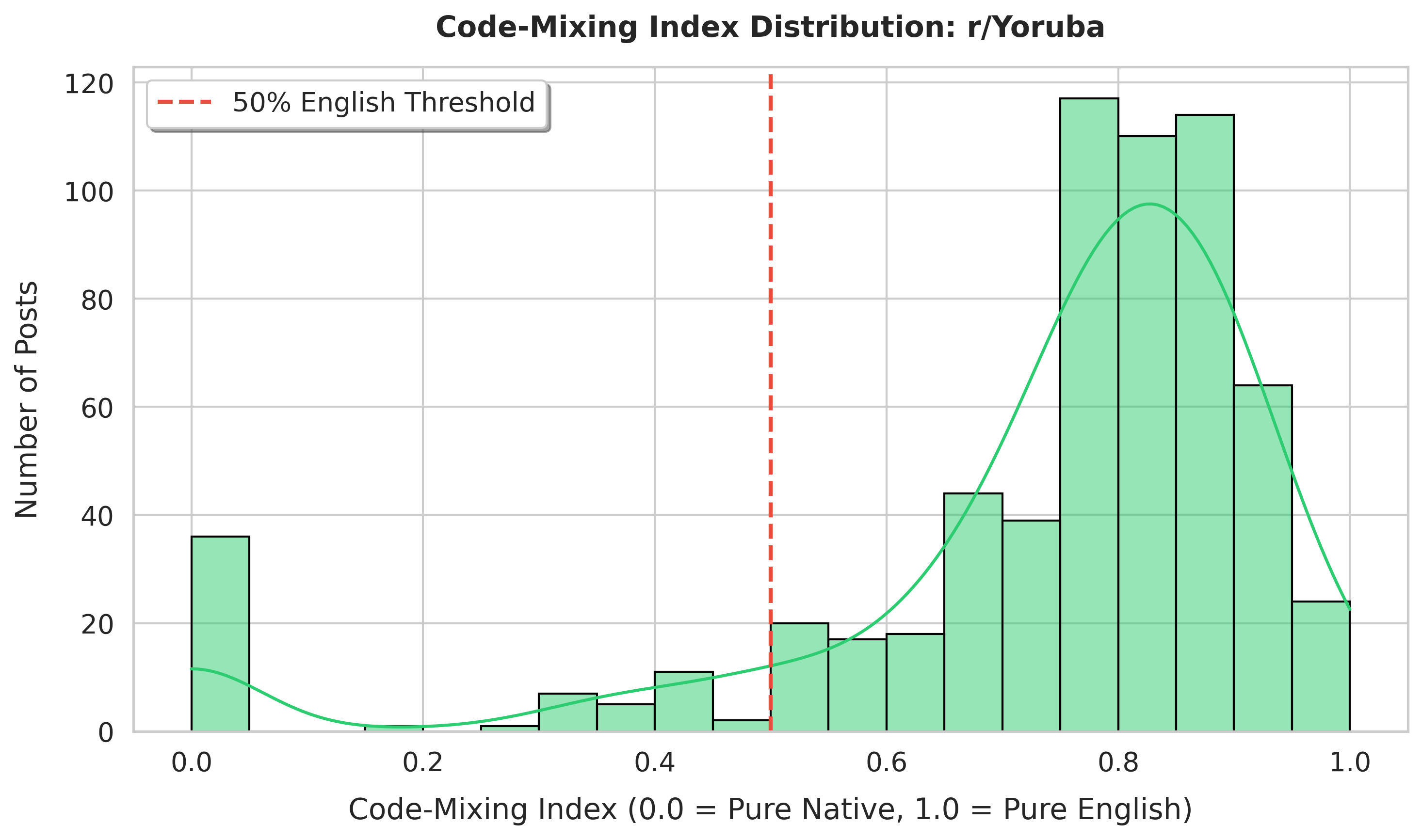}\hfill
    \includegraphics[width=0.48\linewidth]{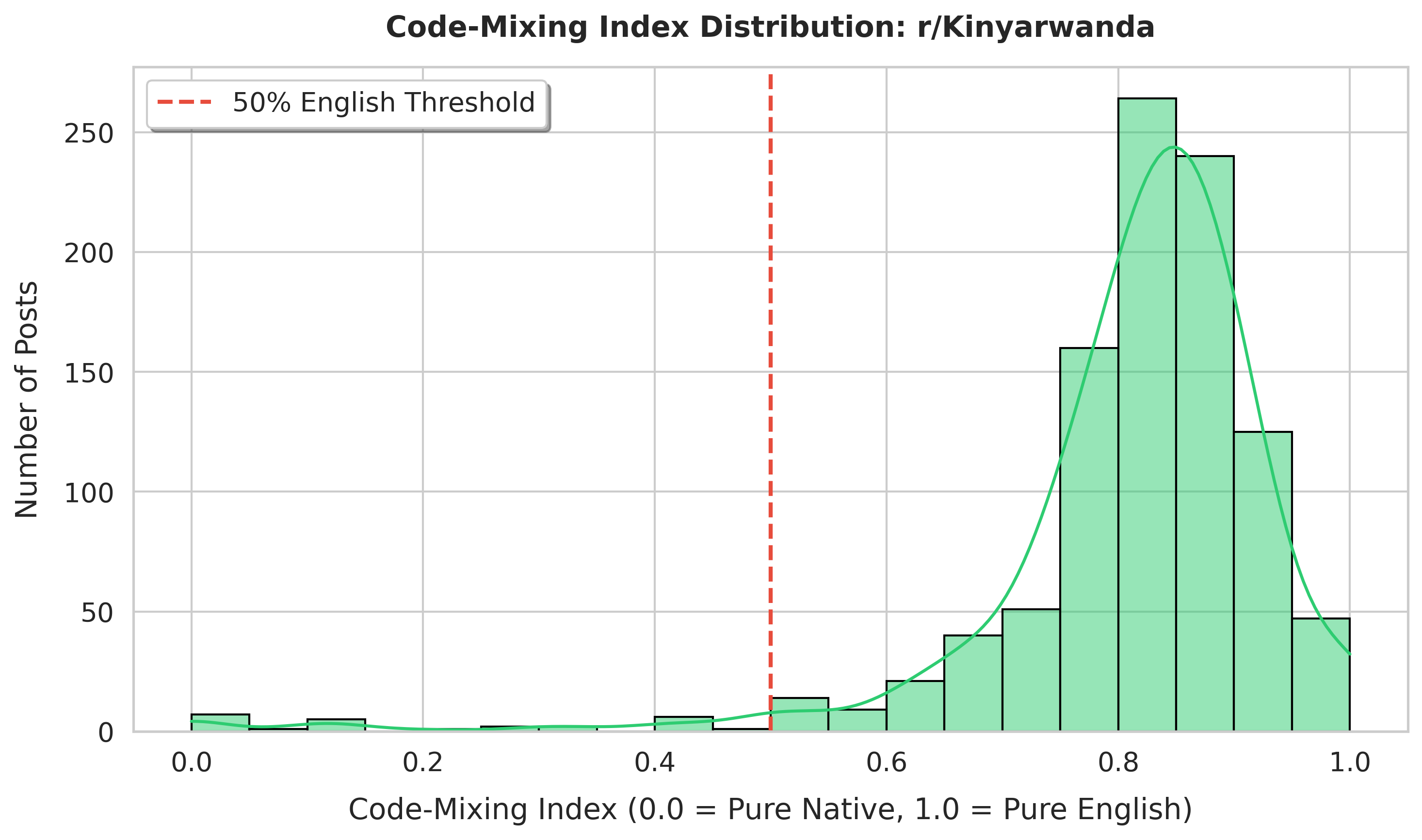}
    \\[0.3cm]
    \includegraphics[width=0.48\linewidth]{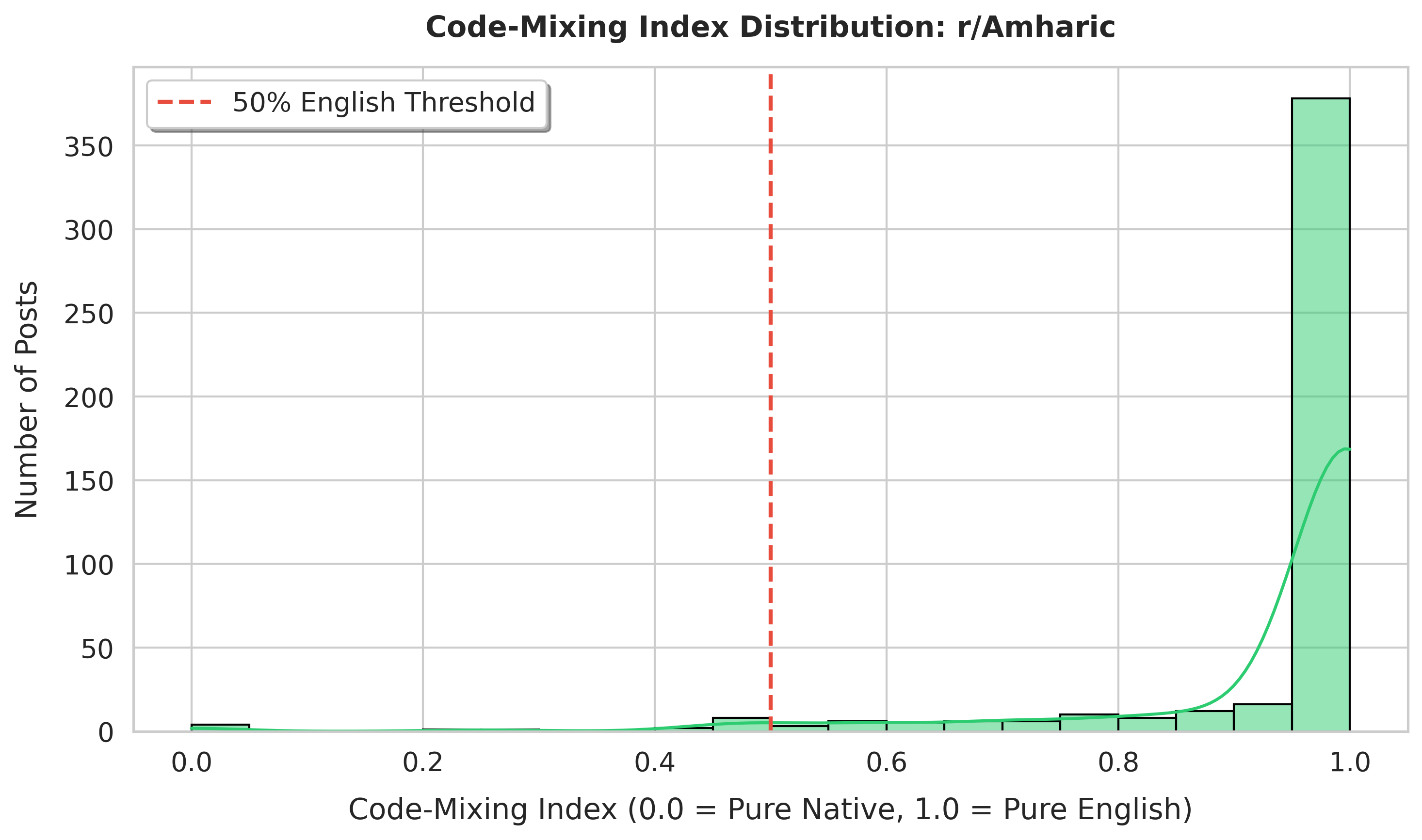}
    \caption{Empirical Distribution of the Code-Mixing Index (CMI) across the three Reddit corpora. The x-axis represents the CMI score, where 0.0 is pure native language and 1.0 is pure English (or exclusively Latin script for Amharic). The y-axis represents the absolute frequency of posts. Notably, all three distributions exhibit a heavy left-skew (a long tail towards 0.0), with the vast majority of the probability mass concentrated between a CMI of $0.5$ and $1.0$. This visualizes that the conversational data is systematically dominated by English code-switching rather than isolated native loanwords.}
    \label{fig:cmi_histograms}
\end{figure}

The resulting distributions (Figure \ref{fig:cmi_histograms}) prove that the Reddit corpora are not native texts: the average CMI for Yoruba, Kinyarwanda, and Amharic was 0.734, 0.812, and 0.942, respectively. Furthermore, 87.1\% of Yoruba posts, 96.7\% of Kinyarwanda posts, and 96.6\% of Amharic posts contained a CMI greater than 0.5 (meaning they were majority English). This empirical formulation explains why AfroLID hallucinated false labels and GlotLID defaulted to English: the underlying text was systematically code-switched beyond the threshold where standard LID architectures can perform reliably.

\subsection{News Topic Classification}
To gain a deeper understanding of the content within our news datasets, we conducted a manual topic classification of news articles from each source. The results indicate that the news media content reflects distinct cultural and societal priorities as seen in Table \ref{tab:news_topics}.

\begin{table}[htbp]
\centering
\caption{Distribution of News Topics by Language Dataset. The values represent the percentage of articles from each structured news source (BBC Yoruba, Kinyarwanda News, Fana Amharic) manually annotated into each topical category. This breakdown highlights the distinct editorial priorities across the different linguistic regions.}
\begin{tabular}{lccc}
\toprule
\textbf{Topic} & \textbf{Yoruba} & \textbf{Kinyarwanda} & \textbf{Amharic} \\
\midrule
Business & 1.61 & 31.43 & 40.70 \\
Education & 6.45 & 0.00 & 1.16 \\
Entertainment & 17.74 & 17.14 & 8.14 \\
Health & 4.84 & 12.38 & 0.00 \\
History & 1.61 & 11.43 & 0.00 \\
Politics & 27.42 & 11.43 & 15.12 \\
Sports & 14.52 & 0.00 & 16.28 \\
Others & 25.81 & 16.19 & 18.60 \\
\bottomrule
\end{tabular}
\label{tab:news_topics}
\end{table}

The analysis reveals distinct content patterns across the three languages:
\begin{itemize}
    \item \textbf{Amharic News}: The overwhelming dominance of business content at 40.7\% suggests a strong focus on economic and commercial affairs in Ethiopian Amharic media. This is complemented by significant coverage of sports (16.28\%) and politics (15.12\%), indicating a well-rounded news ecosystem that balances economic reporting with civic and recreational interests. The complete absence of health and history coverage may reflect editorial priorities that favor immediate economic concerns over longer-term societal issues or historical context.
    \item \textbf{Kinyarwanda News}: While business content leads at 31.43\%, the distribution is more balanced across categories compared to Amharic news. The substantial presence of entertainment (17.14\%) and health coverage (12.38\%) suggests a media landscape that addresses both informational and lifestyle needs of its audience. The equal representation of history and politics (both at 11.43\%) indicates attention to both contemporary governance and cultural heritage, while the complete absence of sports coverage may reflect different recreational priorities or media resource allocation in the Rwandan context.
    \item \textbf{Yoruba News}: The near-equal emphasis on politics (27.42\%) and diverse content categorized as "Others" (25.81\%) reveals a media environment heavily focused on civic engagement and varied community interests. The strong presence of entertainment content (17.74\%) and notable education coverage (6.45\%) - which is absent in the other two languages - suggests a media ecosystem that serves both democratic participation and knowledge dissemination functions. The relatively balanced distribution across most categories indicates a more diversified news landscape compared to the business-heavy focus seen in the other two languages.
\end{itemize}

\noindent
These distinct patterns likely reflect not only different cultural priorities but also varying economic conditions, political climates, and media industry structures within each linguistic community. The data suggests that Amharic media serves primarily economic information needs, Kinyarwanda media balances practical life concerns with civic awareness, and Yoruba media emphasizes democratic participation and educational content.

\subsection{Discussion of Findings}

The comparative analysis of GlotLID, AfroLID, and Llama 3.3 70B sheds light on the complex dynamics of African languages in digital spaces. Our findings reveal two critical, interconnected insights.
\noindent
First, the online conversational platforms, even those clearly dedicated to specific African linguistic communities, are not reliable sources for clean, monolingual data. The high prevalence of English and the pervasive nature of code-switching are a direct reflection of the sociolinguistic realities of many urban African societies, where English often serves as a lingua franca for digital interaction. The data shows that even when members of these communities engage on platforms like Reddit, their interactions are heavily mediated by English. This creates a significant challenge for researchers, as the raw data does not fully represent the authentic, spoken language. Furthermore, the catastrophic failure of standard LID tools on this data should not be viewed as a failure of their internal neural architectures, but rather as a symptom of an extreme distribution shift. These models are predominantly trained on clean, formal text (such as Wikipedia and religious texts), rendering them highly brittle when exposed to the heavily code-switched distributions prevalent on modern social media. It underscores a crucial need to rethink data collection strategies beyond simply scraping text from existing social media platforms.
\noindent
Second, the consistent high performance of both AfroLID and the LLM on news media sources demonstrates a clear path forward. These platforms offer a valuable, high-quality resource for data collection because their content is intentionally created in a cleaner, more formal, and monolingual style. This finding is particularly significant because it empowers African communities to be producers of high-quality digital content in their own languages. By leveraging these existing, professionally curated sources, we can build a robust foundation for language technologies that is more reflective of the languages as they are written and spoken in formal contexts.

\subsection{Ethical Considerations}
This research adheres to ethical guidelines for computational linguistics research using publicly available data collected from Reddit discussions and news media social pages in accordance with platform terms of service.
To protect user privacy, we removed usernames from all collected posts, ensuring individual contributions cannot be directly attributed to specific users. Our analysis focuses on aggregated linguistic patterns, treating the data as representative samples of online language use while maintaining user anonymity.
This methodology enables responsible advancement of African language technologies, contributing to discussions on improved natural language processing capabilities for African languages while respecting online user privacy.

\subsection{Future Considerations}

Based on these findings, we propose several avenues for future research that are specifically tailored to the unique linguistic landscape of Africa.
\begin{itemize}
\item \textbf{Model Development for Code-Switched Data}: Given that code-switching is a natural and integral part of digital communication for many Africans, future efforts must move beyond simply collecting clean data. The next frontier is to develop new models and fine-tune existing ones to better understand and process code-switched data. Instead of treating code-switching as noise, these models should be designed to recognize it as a legitimate and important linguistic phenomenon, thereby creating technologies that are more relevant and useful to the people who will use them.
\item \textbf{Leveraging Audio Data}: Our findings highlight a disconnect between written social media content and the vibrant, multilingual reality of spoken language. A promising avenue for data collection is to look into audio-based social media platforms. By employing strong Automatic Speech Recognition (ASR) technology tailored to African languages, we could transcribe spoken conversations. This approach would allow us to capture more authentic, conversational language use, including the nuances of intonation, rhythm, and natural code-switching that are lost in written text. This would provide a more representative dataset for building truly conversational LLMs.
\item \textbf{Afro-Centric Benchmarking}: To measure true progress, we need to create benchmarks that are culturally and linguistically relevant. These benchmarks should not only test a model's ability to handle monolingual text but also its competence in managing tasks that involve code-switching, transliteration, and context-specific cultural references. By developing and using these benchmarks, we can ensure that language technologies are not just performant, but also deeply respectful and useful for the communities they are intended to serve.
\end{itemize}
\noindent
\section{Conclusion} This study has highlighted a critical divergence in the digital representation of African languages. While local news media provides a valuable source of clean, monolingual data for languages like Yoruba, Kinyarwanda, and Amharic, conversational platforms like Reddit are characterized by code-switching and data scarcity. Our findings demonstrate that current language detection tools, which perform well on formal text, struggle significantly with this informal, mixed-language content. This research underscores a fundamental challenge for AI development in low-resource language communities. We conclude that future research must move beyond the reliance on traditional, clean datasets and instead focus on developing models specifically designed to handle code-switched text, and audio conversations. This will be essential for creating inclusive and effective language technologies that truly reflect the dynamic and diverse linguistic realities of African users in the digital age.
\newpage
\bibliographystyle{unsrt}
\bibliography{ijcai26}

\end{document}